\documentclass[english]{sbrt}
\usepackage{adjustbox}

\usepackage[utf8]{inputenc}

\usepackage{comment}
\usepackage{cite}
\usepackage{multirow}
\usepackage{graphicx}
\usepackage{acronym}
\usepackage{float}
\usepackage{amsmath}
\usepackage{caption}
\usepackage{pgfplots}
\usepackage{xcolor}
\pgfplotsset{compat=1.18}
\usepackage{tikz}
\makeatletter
\usepackage{orcidlink}
\usepackage{adjustbox}    
\usepackage{booktabs}     
\usepackage{array}        

\newcommand{\figurecaptionstyle}{%
  \long\def\@makecaption##1##2{%
    \vskip\abovecaptionskip
    \sbox\@tempboxa{{\small ##1: ##2}}%
    \ifdim \wd\@tempboxa >\hsize
      {\small ##1: ##2\par}%
    \else
      \global \@minipagefalse
      \hb@xt@\hsize{\hfil\box\@tempboxa\hfil}%
    \fi
    \vskip\belowcaptionskip
  }%
}
\newcommand{\tablecaptionstyle}{%
  \long\def\@makecaption##1##2{%
    \vskip\abovecaptionskip
    \sbox\@tempboxa{{\small TABLE~\thetable}}%
    \ifdim \wd\@tempboxa >\hsize
      {\centering \small TABLE~\thetable\par}%
    \else
      \global \@minipagefalse
      \hb@xt@\hsize{\hfil\box\@tempboxa\hfil}%
    \fi
    {\centering \small ##2\par}%
    \vskip 3pt 
  }%
}

\renewcommand{\fnum@table}{\tablename~\thetable}

\AtBeginEnvironment{table}{\tablecaptionstyle}
\AtBeginEnvironment{table*}{\tablecaptionstyle}
\AtBeginEnvironment{figure}{\figurecaptionstyle}
\AtBeginEnvironment{figure*}{\figurecaptionstyle}

\makeatother

\usepackage[english]{babel}
\usepackage[utf8]{inputenc}

\begin{document}

\title{Evaluation of AutoML Frameworks for IDS under Imbalanced Data Conditions of NSL-KDD Dataset}

\author{Wiliane Carolina Silva, Evandro César Vilas Boas~\orcidlink{0000-0002-7225-7783}, and Felipe A. P. de Figueiredo~\orcidlink{0000-0002-2167-7286} \vspace{-0.5cm}
\thanks{Wiliane Carolina Silva, Evandro César Vilas Boas, and Felipe A. P. Figueiredo are with the Cybersecurity and Artificial Intelligence Laboratory (CS\&I Lab) the and Wireless and Artificial Intelligence Laboratory (WAI Lab) of the National Institute of Telecommunications (Inatel), Santa Rita do Sapucaí, Brazil, e-mail: lucas.rosario@get.inatel.br, giovanna.lima@gec.inatel.br, felipe.figueiredo@inatel.br, barbararosa@inatel.br, evandro.cesar@inatel.br. This work was supported by CNPq (311470/2021-1, 403827/2021-3, 306199/2025-4), by the projects XGM-AFCCT-2024-2-5-1, XGM-AFCCT-2024-9-1-1 supported by xGMobile - EMBRAPII-Inatel Competence Center on 5G and 6G Networks, with financial resources from the PPI IoT/Manufatura 4.0 from MCTI (052/2023), signed with EMBRAPII, by RNP, with resources from MCTIC (No. 01245.020548/2021-07), under the Brazil 6G project of the Radiocommunication Reference Center (CRR) of Inatel, Brazil, by FAPEMIG (PPE-00124-23, RED-00194-23, APQ-04523-23, APQ-05305-23, and APQ-03162-24), and by FINEP (nº 1060/2 contract 01.25.0883.00)}%
}

\maketitle

\markboth{XLIV BRAZILIAN SYMPOSIUM ON TELECOMMUNICATIONS AND SIGNAL PROCESSING - SBrT 2026, SEPTEMBER 29TH TO OCTOBER 2ND, 2026, SALVADOR, BA}{}
\acrodef{6G}{Sixth Generation of Mobile Networks}
\acrodef{5G}{Fifth Generation of Mobile Networks}
\acrodef{URLLC}{\textit{Ultra-Reliable and Low-Latency Communications}}
\acrodef{mMTC}{\textit{Massive Machine-Type Communications}}
\acrodef{eMBB}{\textit{Enhanced Mobile Broadband}}
\acrodef{QoS}{\textit{Quality of Service}}
\acrodef{QoE}{\textit{Quality of Experience}}
\acrodef{3GPP}{\textit{3rd Generation Partnership Project}}
\acrodef{NFV}{\textit{Network Function Virtualization}}
\acrodef{SDN}{\textit{Software Defined Networking}}
\acrodef{UE}{\textit{User Equipment}}
\acrodef{RAN}{\textit{Access Network}}
\acrodef{MCC}{\textit{Mobile Country Code}} 
\acrodef{MNC}{\textit{Mobile Network Code}}
\acrodef{DNNs}{\textit{Data Network Names}}
\acrodef{UE}{User}
\acrodef{gNB}{Base Radio Station do 5G}
\acrodef{SST}{\textit{Slice Service Type}}
\acrodef{SD}{\textit{Slice Differentiator}}
\acrodef{S-NSSAI}{\textit{Single Network Slice Selection Assistance Information}}
\acrodef{ICMP}{\textit{Internet Control Message Protocol}}
\acrodef{SLAs}{\textit{Service Level Agreements}}
\acrodef{5QI}{5G QoS \textit{Identifier}}
\acrodef{ARP}{\textit{Allocation and Retention Priority}}
\acrodef{GBR}{\textit{Guaranteed Bit Rate}}
\acrodef{non-GBR}{\textit{Non-Guaranteed Bit Rate}}
\acrodef{ITU-T}{\textit{International Telecommunication Union}}
\acrodef{RTT}{\textit{Round-Trip Time}}
\acrodef{PDV}{\textit{Packet Delay Variation}}
\acrodef{PLR}{\textit{Packet Loss Rate}}
\acrodef{AMF}{\textit{Access and Mobility Management Function}}
\acrodef{SMF}{\textit{Session Management Function}}
\acrodef{UPF}{\textit{User Plane Function}}
\acrodef{NRF}{\textit{Network Repository Function}}
\acrodef{NSSF}{\textit{Network Slice Selection Function}}
\acrodef{DNNs}{\textit{Data Network Names}} 

\begin{abstract}
This work investigates the impact of severe class imbalance on the performance of automated machine learning (AutoML) frameworks for multiclass network intrusion detection using the NSL-KDD dataset. Unlike previous studies that simplify the problem through binary classification or minority-class removal, we preserve the original five-class distribution, including highly underrepresented attacks such as R2L and U2R, enabling a realistic evaluation of imbalance-sensitive learning behavior. Nine open-source AutoML frameworks were analyzed under a unified and reproducible experimental protocol, considering differences in architectural design, ensemble strategies, validation procedures, hyperparameter optimization, and imbalance-handling mechanisms. The results demonstrate that frameworks incorporating ensemble learning and imbalance-aware optimization achieve better minority-class discrimination. PyCaret obtained the best overall performance, reaching 66\% macro-F1, followed by AutoGluon with 55\%, whereas frameworks lacking native balancing support exhibited significant degradation in minority-class detection capability. The analysis further shows that accuracy-oriented optimization alone is insufficient for highly imbalanced IDS scenarios, since high-weighted metrics may coexist with poor generalization on rare attack categories. As a contribution, this work establishes a standardized benchmark for AutoML-based intrusion detection under severe multiclass imbalance, highlighting current architectural limitations and the need for native integration of imbalance-aware optimization, resampling, and stratified evaluation strategies into automated learning pipelines. The source code is publicly available.\footnote{\url{https://github.com/wilicarol/Code-PyCaret-TCC.git}}
\end{abstract}

\section{Introduction} \label{introdução}

Network security is a critical domain within cybersecurity due to the exponential growth in data exchange among interoperable systems, increasing exposure to cyber threats. In this context, Network Intrusion Detection Systems (NIDS) are employed to monitor network traffic and identify malicious activities through signature-based or anomalous behavioral analysis~\cite{Wiliane_06.pdf}. However, traditional signature-based NIDS are limited in detecting novel threats, such as zero-day attacks, and handling increasingly complex traffic patterns. To overcome these limitations, machine learning (ML) techniques have been adopted in anomaly-based NIDS, enabling the identification of deviations from normal network behavior that may indicate malicious activity~\cite{17,Wiliane_06.pdf,21, Wiliane_08.pdf}. Nevertheless, ML-based NIDS rely on multiclass datasets with severe class imbalance, in which minority attack categories are underrepresented due to the lower frequency of rare or customized attacks, biasing the learning process toward majority classes with low detection performance for minority attacks.

Although ML-based classification models offer significant advantages, no single algorithm consistently outperforms others across all scenarios due to variations in dataset characteristics and attack profiles~\cite{Wiliane_06.pdf,21}. To address this variability, automated machine learning (AutoML) automates key stages of the ML pipeline, including preprocessing, algorithm selection, hyperparameter optimization, and ensemble construction, enabling the systematic evaluation of multiple configurations under a unified protocol~\cite{qi2021autogluon, sarangpure2023automating,olson2016tpot,ledell2020h2o,feurer2022auto,wang2021flaml,zimmer2021auto,jin2019auto}. However, in multiclass imbalanced scenarios, effective model selection requires evaluation metrics that prioritize minority classes, such as the $F_{1\text{-score}}$, since accuracy-oriented optimization may favor models with poor recall for rare classes~\cite{Marcelo}.

\begin{figure*}[!t]
\centering
\begin{adjustbox}{width=\textwidth}
\includegraphics{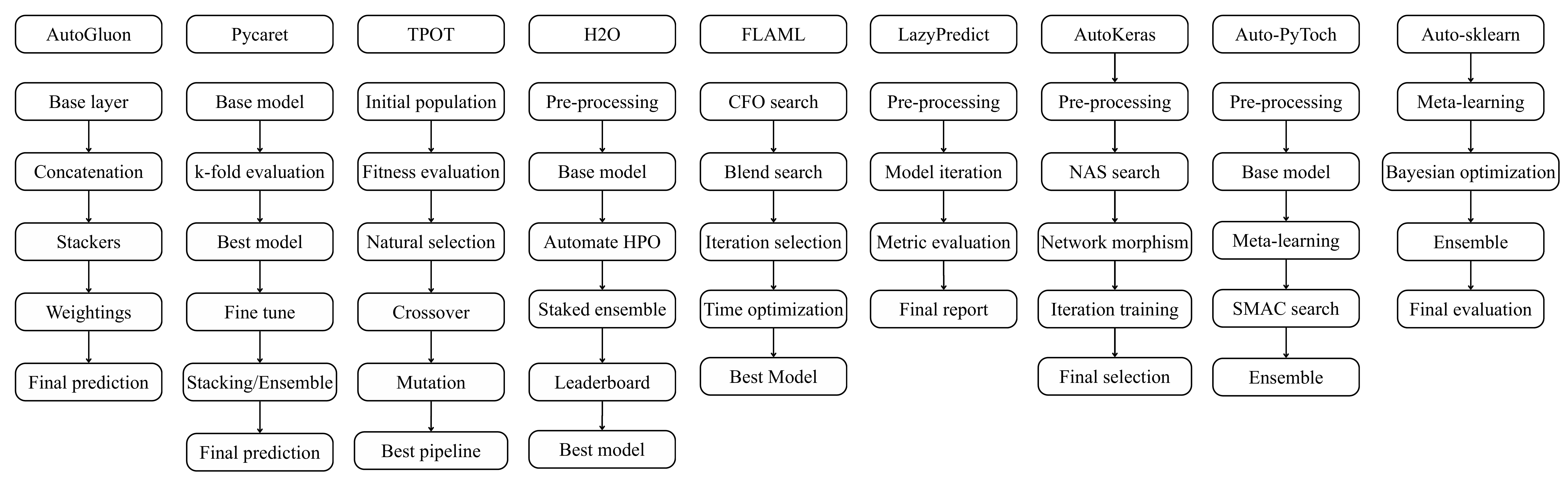}
\end{adjustbox}
\caption{AutoML frameworks pipeline: Autogluon, PyCaret, TPOT, H2O, FLAML, LazyPredict, AutoKeras, Auto-PyTorch and Auto-sklearn.}
\label{fig:arquiteturas}
\end{figure*}

Concerning the NIDS domain, the NSL-KDD dataset~\cite{NSL-KDD1} is a widely adopted benchmark for evaluating ML-based intrusion detection models. Nevertheless, NSL-KDD presents severe class imbalance, motivating studies focused on improving both overall accuracy and minority-class $F_{1\text{-score}}$. Related works on the NSL-KDD dataset address both binary and multiclass classification problems~\cite{17,Wiliane_06.pdf,21,9,Wiliane_05.pdf, Wiliane_08.pdf}. These approaches often follow a similar manual pipeline: conversion of categorical attributes into numerical representations, normalization or standardization, optional feature selection or dimensionality reduction, and the selection of a classifier with tuned parameters. Despite advances in ML, deep learning (DL)~\cite{9,Wiliane_05.pdf}, and hybrid approaches, most studies rely on manually designed workflows and heterogeneous evaluation protocols, limiting reproducibility and comparability. In this context, AutoML remains underexplored for IDS, although it offers standardized experimentation, automated hyperparameter optimization, and native integration of balancing techniques~\cite{7}.

Based on that, this work applies different AutoML architectures to conduct a comparative performance analysis on the NSL-KDD dataset, aiming to assess the effectiveness of these approaches in imbalanced multiclass scenarios. To support this investigation, nine open-source frameworks were selected: AutoGluon~\footnote{\url{https://auto.gluon.ai/stable/index.html}}, PyCaret~\footnote{\url{https://pycaret.gitbook.io/docs/}}, TPOT~\footnote{\url{https://epistasislab.github.io/tpot/latest/}}, H2O~\footnote{\url{https://docs.h2o.ai/h2o/latest-stable/h2o-docs/automl.html}}, Auto-Sklearn~\footnote{\url{https://automl.github.io/auto-sklearn/master/}}, FLAML~\cite{FLAML1}, LazyPredict~\footnote{\url{https://github.com/shankarpandala/lazypredict}}, Auto-PyTorch~\footnote{\url{https://github.com/automl/Auto-PyTorch}} and AutoKeras~\footnote{\url{https://autokeras.com/}}. Despite sharing the common objective of automating ML pipelines, these frameworks differ in terms of internal methodology, supported data modalities, and optimization strategies. Consequently, they may recommend distinct ML algorithms and produce either similar or different performance outcomes.

This work contributes to advances in ML-based IDS by evaluating the impact of severe class imbalance in the NSL-KDD dataset on the performance of different AutoML frameworks. Furthermore, it establishes a standardized and reproducible benchmark, highlighting the need for more robust pipelines and imbalance-handling strategies in IDS applications. The paper is organized into five sections. Section~\ref{sec:frameworks_autoML} describes the internal architectures of each AutoML. Section~\ref{nslkdd} presents the NSL KDD dataset. Section~\ref{sec:resultados_comparativos} discusses the results and compares them with related works. Finally, Section~\ref{conclusio} summarizes the main conclusions and contributions.

\section{Discussion of AutoML Frameworks}
\label{sec:frameworks_autoML}

This section presents the AutoML frameworks, focusing on their training architectures, optimization strategies, pipeline construction, and limitations related to the imbalanced NSL-KDD dataset. Although methodologically distinct, the frameworks follow a similar workflow involving data loading, train-validation splitting, and iterative model evaluation guided by performance metrics. The analyses are based on the training workflows illustrated in Figure~\ref{fig:arquiteturas}.

\subsection{AutoML Framework features}

AutoGluon automates ML model development for tabular, image, and text data~\cite{qi2021autogluon}. Its pipeline integrates preprocessing and weighted ensembles of heterogeneous models, including LightGBM, CatBoost, XGBoost, neural networks, and Random Forest. It automatically performs train-validation splitting. In this study, \texttt{holdout-frac=0.2} was adopted. Although AutoGluon does not provide native resampling mechanisms, its weighting strategy may favor models with better minority-class performance according to metrics such as ROC-AUC and $F_{1\text{-score}}$. However, the absence of explicit balancing techniques may bias predictions toward majority classes in NSL-KDD.

PyCaret is a low-code AutoML library for Python that automates supervised learning tasks such as preprocessing, estimator comparison, hyperparameter optimization with Optuna, and ensemble construction~\cite{sarangpure2023automating}. The framework supports models including KNN, Random Forest, LightGBM, XGBoost, and Support Vector Machine. It addresses class imbalance through explicit resampling methods using the \texttt{fix\_imbalance} parameter, including the synthetic minority over-sampling technique(SMOTE), although class weighting must be configured manually when supported by the model. In this study, the default 70\%/30\% train-validation split was adopted, and the macro $F_\text{1}$ metric was added through the \texttt{add\_metric} function to improve minority-class evaluation.

Tree-based Pipeline Optimization Tool (TPOT) is based on scikit-learn and uses genetic programming (GP) to optimize complete learning pipelines~\cite{olson2016tpot}. Its architecture represents pipelines as graphs, enabling combinations of transformations, feature-selection methods, and classifiers. The evolutionary process evaluates a population of pipelines through cross-validation (CV), applying genetic operators such as mutation and crossover to generate new populations. Evaluation is performed on the training set using internal CV, without a fixed holdout partition. Concerning imbalance handling, TPOT is inherently agnostic, lacking automatic resampling or weighting mechanisms~\cite{olson2016tpot}. Consequently, it tends to converge toward solutions favoring majority classes when guided by imbalance-insensitive metrics.

H2O AutoML is a scalable platform developed by \textit{H2O.ai}, designed for large-scale model construction. Its architecture trains multiple algorithms in parallel, while also automatically building stacked ensembles. Preprocessing includes categorical-variable encoding through target encoding~\cite{ledell2020h2o}. In this work, the split-frame function was employed with an 80\% training and 20\% validation ratio. H2O offers automatic balancing mechanisms through the \texttt{balance\_classes} parameter, which applies minority-class oversampling and also allows weighting within the loss function. Despite these functionalities, the intensive use of ensembles may propagate bias inherited from base learners, mainly under severe imbalance conditions such as those found in NSL-KDD, when automatic balancing is insufficient.

\begin{table}[t]
\centering
\caption{Mechanisms for handling imbalances in AutoML frameworks.}
\label{tab:desbalanceamento_frameworks}
\resizebox{8cm}{!}{
\begin{tabular}{|c|c|c|c|}
\hline
\textbf{Framework} & \textbf{Native support} & \textbf{Mechanism} & \textbf{Activation} \\
\hline
\centering AutoGluon      & Yes  & Ensemble weighing          & Automatic \\
\hline
\centering H2O AutoML     & Yes  & Internal resampling/Class weighting                     & Automatic \\
\hline
\centering PyCaret        & Partial & SMOTE/metric weighting           & Manual \\
\hline
\centering Auto-sklearn   & Partial & Class weighting               & Manual \\
\hline
\centering Auto-PyTorch   & Yes  & Sample weight/ Resampling (fold)                  & Manual \\
\hline
\centering FLAML          & Partial & Sample weighting                 & Manual \\
\hline
\centering TPOT           & Partial & SMOTE       & Manual \\
\hline
\centering LazyPredict    & No  & It does not offer support                                      & - \\
\hline
\centering AutoKeras      & Partial & Class weighting                           & Manual \\
\hline
\end{tabular}
}
\end{table}

Auto-sklearn is a scikit-learn-based toolkit that automates algorithm selection and hyperparameter optimization through meta-learning and Bayesian optimization, implemented using sequential model-based algorithm configuration (SMAC)~\cite{feurer2022auto}. It integrates meta-learning for efficient search initialization, followed by the construction of automatic ensembles. Internal evaluation is performed through a CV. Herein, Auto-sklearn was configured to optimize the macro $F_\text{1}$ metric through \texttt{metric=autosklearn.metrics.f1\_macro}. To address the imbalance, Auto-sklearn supports class-weight-based strategies, although it does not provide native resampling mechanisms. Due to its dependence on meta-learning, limitations may emerge in specialized domains such as network security, where NSL-KDD presents highly specific characteristics.

Fast Lightweight AutoML (FLAML) is focused on computational efficiency. Its architecture implements the cost-frugal optimization (CFO) strategy, which seeks to minimize computational cost during optimization~\cite{wang2021flaml}. The search process is iterative and prioritizes low-cost configurations, rapidly discarding those that fail to demonstrate promising performance.
In this study, FLAML was configured using the macro $F_\text{1}$ metric for optimization, \texttt{split\_ratio=0.2} to allocate 20\% of the data for validation, and \texttt{ensemble=True} to enable ensemble learning. FLAML also supports class weighting through the \texttt{sample\_weight} argument. However, it lacks automatic resampling mechanisms. Its early-discard strategy may eliminate computationally expensive configurations that could potentially achieve superior performance on rare classes.

LazyPredict is an exploratory benchmarking tool that executes multiple scikit-learn models using default hyperparameters. It neither performs hyperparameter optimization nor feature engineering. Herein, the data is previously divided into 80\% training and 20\% validation subsets. LazyPredict is agnostic to class imbalance, providing neither weighting nor automatic resampling support, which makes it highly susceptible to bias toward majority classes. Auto-PyTorch is a PyTorch-based toolkit aimed at jointly optimizing architectures and hyperparameters for DL and tabular-data tasks~\cite{zimmer2021auto}. Its architecture combines Bayesian optimization, meta-learning, and multifidelity strategies. In this study, the split ratio was defined as 80\% for training and 20\% for validation. The framework supports class weighting within the loss function and may incorporate resampling techniques into the pipeline. However, the high computational cost associated with architecture search may limit its efficiency for large datasets, especially under severe imbalance conditions such as NSL-KDD.

AutoKeras is an AutoML framework specialized for DL, built upon Keras and TensorFlow~\cite{jin2019auto}. Its architecture employs neural architecture search (NAS) guided by Bayesian optimization to explore neural-network architectures. By default, the framework uses the last 20\% of the training data as a validation set. It supports class weighting and the integration of resampling techniques when appropriately configured. However, due to the computationally intensive nature of architecture search, optimization may become concentrated on minimizing global loss, disregarding subtle minority-class patterns unless imbalance-sensitive metrics are explicitly defined. Overall, despite sharing common stages such as data splitting and iterative evaluation, the frameworks differ substantially in their imbalance-handling capabilities. Table~\ref{tab:desbalanceamento_frameworks} summarizes the mechanisms identified in the analyzed frameworks.

\section{NSL KDD dataset}
\label{nslkdd}
The NSL-KDD dataset contains network traffic records that simulate realistic attack patterns. Each NSL-KDD record is labeled either as normal traffic or as belonging to one of four attack categories: denial of service (DoS), Probe, remote to local (R2L), and user to root (U2R)~\cite{NSL-KDD1}. Each instance contains 41 input attributes and an associated class label. Class imbalance constitutes one of the main challenges of this dataset, providing a realistic benchmark for evaluating ML algorithms, particularly in multiclass classification tasks~\cite{10}. The NSL-KDD dataset provides 125,972 training samples and two testing subsets: \texttt{KDDTest+}, containing 22,544 samples, and \texttt{KDDTest-21}, containing 11,850 samples. In this study, the \texttt{KDDTest+} subset was employed to evaluate the performance of the trained models. Table~\ref{tab:distribuicao_nslkdd} presents the class-wise sample counts and proportions.

\begin{table}[!ht]
\centering
\caption{Distribution of samples, percentage by class, and imbalance ratio in the KDDTrain$+$ and KDDTest$+$ sets.}
\label{tab:distribuicao_nslkdd}
\begin{adjustbox}{width=9cm}
\begin{tabular}{|c|c|c|c|c|c|c|}
\hline
\multirow{3}{*}{\textbf{Classes}} & \multicolumn{3}{c|}{\textbf{KDDTrain$+$}} & \multicolumn{3}{c|}{\textbf{KDDTest+}} \\
\cline{2-7}
& \textbf{Sample} & \textbf{Percentage (\%)} & \textbf{IR} & \textbf{Sample} & \textbf{Percentage (\%)} & \textbf{IR}\\
\hline
Normal & 67,343 & 53.45 & 1.00& 9,711 & 43.07&1.00 \\\hline
DoS & 45,927 & 36.45 & 1.47 & 7,458 & 33.08 & 1.30 \\\hline
Probe & 11,656 & 9.25 & 5.78 & 2,421 & 10.74& 4.01 \\\hline
R2L & 995 & 0.79 & 67.68 & 2,754 & 12.22 & 3.53 \\\hline
U2R & 52 & 0.06 & 1295.06 & 200 & 0.89 & 48.55\\\hline
\end{tabular}
\end{adjustbox}
\end{table}

\begin{table*}[!ht] 
\centering
\caption{Comparison between AutoML frameworks and related work using NSL-KDD to handle five classification tasks.}
\label{tab:comparativo_nslkdd}
\begin{adjustbox}{width=\textwidth}
\begin{tabular}{|c|c|c|c|c|c|c|c|}
\hline
\textbf{Reference}    & \textbf{Framework} & \textbf{Model}                 & \textbf{$F_1$-weighted} & \textbf{$F_1$-Macro} & \textbf{Accuracy} & \textbf{Precision} & \textbf{Recall}\\
\hline
This work             & PyCaret            & CatBoost Classifier            & 80.00\% & 66.00\% & 82.00\% & 85.00\% & 82.00\%  \\
This work             & AutoGluon          & WeightedEnsemble               & 74.01\% & 55.00\% & 78.00\% & 83.00\% & 78.00\% \\
This work             & AutoKeras          & Multilayer Perceptron          & 72.97\% & 53.65\% & 76.54\% & 80.00\% & 77.00\% \\
This work             & H2O AutoML         & XGBoost                        & 72.00\% & 52.00\% & 76.00\% & 82.00\% & 76.00\% \\
This work             & LazyPredict        & XGBoost                        & 73.00\% & 52.00\% & 77.00\% & 82.00\% & 77.00\% \\
This work             & Auto-Sklearn       & RandomForest + GradientBoosting& 73.00\% & 51.00\% & 77.00\% & 82.00\% & 77.00\% \\
This work             & TPOT               & LGBM Classifier                & 68.00\% & 49.00\% & 71.00\% & 71.00\% & 71.00\% \\
This work             & Auto-PyTorch       & StackingClassifier             & 72.00\% & 49.00\% & 76.00\% & 82.00\% & 76.00\% \\
This work             & FLAML              & StackingClassifier             & 69.00\% & 48.00\% & 73.00\% & 82.00\% & 73.00\% \\
\hline
\cite{Wiliane_06.pdf} & Handcrafted          & Random Forest                  & 89.71\% &71.10\%  & 90.14\% & 92.00\% & 90.00\%  \\
\cite{Wiliane_06.pdf} & Handcrafted          & Gradient Boost                 & 89.53\% & 68.99\%  & 90.45\% & 91.00\% & 90.00\% \\
\cite{21}             & Handcrafted          & Voting Ensemble                & 84.90\% & 69.83\%  & 85.20\% & 86.50\% & 85.20\% \\
\cite{9} $\bullet$    & Handcrafted          & SSDDQN                         & 76.22\% & -- & 79.43\%  & 82.81\% & 79.43\%\\
\cite{Wiliane_05.pdf} & Handcrafted          & Deep Learning                  & 76.47\% & 59.34\%  & 79.74\% & 82.22& 79.74\%  \\
\hline
\end{tabular}
\end{adjustbox}
\begin{flushleft}
\footnotesize
($\bullet$) indicate that the authors did not specify the test subset employed (KDDTest+ or KDDTest21).\\

\end{flushleft}
\end{table*}

The percentages presented in Table~\ref{tab:distribuicao_nslkdd} show that R2L is rare in the training set (0.79\%) while becoming relatively common in the testing set (12.22\%). This difference hinders generalization, since the model is exposed to very few R2L instances during training but is later evaluated against a larger number of such samples. A common consequence is high overall accuracy followed by low recall for minority classes. Under these conditions, cross-validation procedures should be stratified and repeated in order to reduce variance and ensure minimal class representation within each fold. To quantify the class sample difference, the imbalance ratio (IR) of each class \(i\) is adopted:
\begin{equation}
\mathrm{IR}_i = \frac{n_{\max}}{n_i},   
\end{equation}
where \(n_{\max}\) corresponds to the sample count of the majority class and \(n_i\) represents the number of instances in class \(i\). In case \(\mathrm{IR}_i = 1\), class \(i\) has the same frequency as the majority class, and no imbalance exists for that class. As \(\mathrm{IR}_i\) increases, the asymmetry among classes becomes more pronounced, making learning and generalization for class \(i\) difficult and reducing metrics such as recall and \(F_{1\text{-score},i}\)~\cite{IR}. Table~\ref{tab:distribuicao_nslkdd} presents the calculated imbalance ratios for \texttt{KDDTrain$+$} and \texttt{KDDTest+}. An extreme difference can be observed during training for U2R, with \(\mathrm{IR} \approx 1295.06\), and for R2L, with \(\mathrm{IR} \approx 67.68\). In the testing set, U2R remains highly underrepresented, with \(\mathrm{IR} \approx 48.55\), whereas R2L becomes relatively more frequent, with \(\mathrm{IR} \approx 3.53\), characterizing a distribution shift between training and testing.

\section{Results and Discussion}
\label{sec:resultados_comparativos}

This section presents and discusses the results obtained from applying the AutoML frameworks to the NSL-KDD dataset. The experiments involved the frameworks described in Section~\ref{sec:frameworks_autoML}, executed according to their respective official documentation in order to ensure comparability among the obtained results. Table~\ref{tab:comparativo_nslkdd} presents the performance metrics achieved by each AutoML architecture and compares them with related works that proposed handcrafted architectures to deal with the five classification tasks related to the NSL-KDD~\cite{Wiliane_06.pdf,21,9,Wiliane_05.pdf}. Herein, related works that propose binary or four-class classification were excluded from a comparison, since their reported performance metrics stem from practices that compromise realistic evaluation.

The results demonstrate that the AutoML frameworks exhibit different behaviors, which are influenced by their architectural strategies for handling the imbalance inherent to NSL-KDD. PyCaret and AutoGluon achieved the best performances, reaching \(F_{1\text{-macro}}\) values of 66\% and 55\%, respectively. This performance can be attributed to architectures capable of exploring broader combinations of models and hyperparameters. PyCaret, through its modular design based on stacking, ensembles, and refined hyperparameter optimization, efficiently navigates the model search space. In contrast, AutoGluon distinguishes itself through the use of ensemble weighting, hierarchically combining multiple models, which helps mitigate majority-class dominance and provides greater predictive stability, although its \(F_{1\text{-macro}}\) remains below that of PyCaret. The ensemble provided by AutoGluon combines three models: NeuralNetFastAI, Extra Trees with Gini criteria, and Light Gradient Boosting (LGBM).

In contrast, TPOT and FLAML achieved \(F_{1\text{-macro}}\) values of 49\% and 48\%, respectively. These limitations stem from the absence of automatic balancing strategies, limited emphasis on ensembles, and a tendency to prioritize overall accuracy. TPOT, based on genetic programming (GP), favors solutions that maximize global accuracy without necessarily capturing minority-class patterns. Similarly, FLAML prioritizes lightweight and computationally efficient optimization, which reduces its ability to explore more complex models under imbalanced conditions. The FLAML proposed stacking classifier considered the ML models eXtreme Gradient Boosting (XGBoost), LGBM, Random Forest (RF), and Extra Trees (ET). Although Auto-sklearn and Auto-PyTorch employ meta-learning and ensemble strategies, they did not outperform the two best-performing frameworks, indicating that the presence of Bayesian optimization and stacking alone is insufficient. Herein, the Auto-PyTorch stacking classifier included RF, ET, LGBM, and CatBoost.

The related works summarized in Table~\ref{tab:comparativo_nslkdd} achieved their results through specialized preprocessing, balancing, and optimization strategies tailored to the NSL-KDD dataset. The authors from \cite{Wiliane_06.pdf, 21} explored supervised and ensemble-based IDS architectures using feature selection, normalization, hyperparameter tuning, adaptive voting, undersampling, and class-aware weighting to improve multiclass detection, mainly for minority classes such as U2R and R2L. In contrast, \cite{9,Wiliane_05.pdf} focused on DL-based approaches, including reinforcement learning, AutoEncoder-based latent representations, clustering, and dimensionality reduction to improve generalization and reduce computational complexity.

Overall, handcrafted IDS architectures~\cite{Wiliane_06.pdf,21,9} consistently outperformed the default-configured AutoML frameworks under imbalance-sensitive metrics, especially macro $F_1$ and Recall. The manually optimized Gradient Boosting, Random Forest, and adaptive Voting Ensemble models achieved Macro-F1 scores between 68.99\% and 71.10\%, with Recall values up to 90.00\%, demonstrating superior discrimination of minority attack classes. Among the AutoML approaches, PyCaret achieved the best performance, reaching 66.00\% macro $F_1$  and 82.00\% Recall, while most remaining frameworks remained below 55\% macro $F_1$, indicating limited sensitivity to low-frequency attacks.

This difference is mainly associated with the imbalance-aware mechanisms explicitly incorporated into handcrafted approaches, including feature-selection techniques, adaptive weighting, undersampling, latent-space learning, and reinforcement optimization. The AutoML frameworks relied on generalized pipelines and default search spaces without dataset-specific adaptation, limiting their effectiveness in minority-class detection. Nevertheless, despite the absence of manual tuning, feature engineering, or balancing optimization, AutoML frameworks such as PyCaret, AutoGluon, AutoKeras, and H2O AutoML maintained competitive recall and weighted $F_1$  values, demonstrating that modern AutoML solutions can provide robust and reproducible IDS baselines while reducing development complexity and engineering effort.

\section{Conclusion}
\label{conclusio} 

This study presented a systematic evaluation of nine AutoML frameworks applied to the imbalanced five-class NSL-KDD dataset to assess their effectiveness in NIDS scenarios. The results revealed substantial architectural differences, with PyCaret and AutoGluon achieving the best macro $F_{1\text{-score}}$, reaching 66\% and 55\%, respectively, mainly due to mechanisms such as stacking, Optuna-based hyperparameter optimization, and weighted ensembles. In contrast, frameworks without native balancing support, including LazyPredict, TPOT, and FLAML, showed limited capability in detecting minority classes such as R2L and U2R. The findings indicate that AutoML architectures incorporating imbalance-sensitive metrics and ensemble-based strategies provide greater robustness and generalization under multiclass imbalance conditions, with macro $F_1$ proving to be the most appropriate evaluation metric. Furthermore, this work establishes a standardized and reproducible benchmark for AutoML-based IDS evaluation on NSL-KDD, highlighting minority-class handling as the main limitation of current automated solutions. Future research should prioritize the native integration of balancing strategies into search and optimization procedures, including automated oversampling, class-weighting, threshold adjustment, and stratified validation, as well as the exploration of optimized DL architectures and more heterogeneous datasets to improve IDS adaptability and robustness.

\bibliographystyle{IEEEtran}
\bibliography{referencias}
\end{document}